\pdfoutput=1

\documentclass[11pt]{article}

\usepackage{ACL2023}

\usepackage{times}
\usepackage{latexsym}
\usepackage{booktabs}
\usepackage{amsmath}
\usepackage{ulem}
\usepackage[T1]{fontenc}
\usepackage{algorithm}
\usepackage{algorithmic}
\usepackage{enumitem}
\usepackage{graphicx} 
\usepackage{bm}
\usepackage[utf8]{inputenc}

\usepackage{microtype}

\newcommand{\etc}{\textit{etc.}}

\usepackage{inconsolata}

\title{STAR: Constraint LoRA with Dynamic Active Learning for Data-Efficient Fine-Tuning of Large Language Models}

\author{Linhai Zhang$^{\spadesuit}$$^{\diamondsuit}$\thanks{~~Equal Contribution.} \hspace{1.5mm}
Jialong Wu$^{\spadesuit}$$^{\diamondsuit*}$ \hspace{1.5mm}
Deyu Zhou$^{\spadesuit}$$^{\diamondsuit}$\thanks{~~Corresponding Author.} \hspace{1.5mm}
Guoqiang Xu$^{\heartsuit}$\\
$^{\spadesuit}$School of Computer Science and Engineering, Southeast University, Nanjing, China \\
$^{\diamondsuit}$Key Laboratory of New Generation Artificial Intelligence Technology and Its \\
Interdisciplinary Applications (Southeast University), Ministry of Education, China \\ 
$^{\heartsuit}$SANY Group Co., Ltd. \\
\texttt{\{lzhang472, jialongwu, d.zhou\}@seu.edu.cn} \\
\texttt{xuguoqiang-2012@hotmail.com}
} 

\begin{document}
\maketitle

\begin{abstract}
Though Large Language Models (LLMs) have demonstrated the powerful capabilities of few-shot learning through prompting methods, supervised training is still necessary for complex reasoning tasks. 
Because of their extensive parameters and memory consumption, both Parameter-Efficient Fine-Tuning (PEFT) methods and Memory-Efficient Fine-Tuning methods have been proposed for LLMs. 
Nevertheless, the issue of large annotated data consumption, the aim of Data-Efficient Fine-Tuning, remains unexplored. 
One obvious way is to combine the PEFT method with active learning. 
However, the experimental results show that such a combination is not trivial and yields inferior results. Through probe experiments, such observation might be explained by two main reasons: uncertainty gap and poor model calibration.
Therefore, in this paper, we propose a novel approach to effectively integrate uncertainty-based active learning and Low-Rank Adaptation (LoRA).
Specifically, for the uncertainty gap, we introduce a dynamic uncertainty measurement that combines the uncertainty of the base model and the uncertainty of the full model during the iteration of active learning.
For poor model calibration, we incorporate the regularization method during LoRA training to keep the model from being over-confident, and the Monte-Carlo dropout mechanism is employed to enhance the uncertainty estimation.
Experimental results show that the proposed approach outperforms existing baseline models on three complex reasoning tasks.\footnote{ Our code
and results will be available at \url{https://github.com/callanwu/STAR}.
}
\end{abstract}

\section{Introduction}

Large Language Models (LLMs)~\cite{brown2020language,wei2021finetuned,chatgpt,touvron2023llama,llama2,LLMSurvey} have demonstrated the powerful capabilities of zero/few-shot learning with prompting techniques, including In-Context Learning~\cite{dong2022survey} and Chain-of-Thought~\cite{wei2022chain}, where no parameter update is required. 
However, previous studies~\cite{mmlu,yuan2023scaling,bai2023constituency,isik2024scaling} have shown that further fine-tuning is still crucial for tasks involving complex reasoning such as arithmetic reasoning~\cite{roy2016solving,cobbe2021gsm8k} and commonsense reasoning~\cite{mihaylov-etal-2018-suit,clark-etal-2019-boolq}, $\etc$

\begin{figure}[t]
    \centering
    \includegraphics[width=0.49\textwidth]{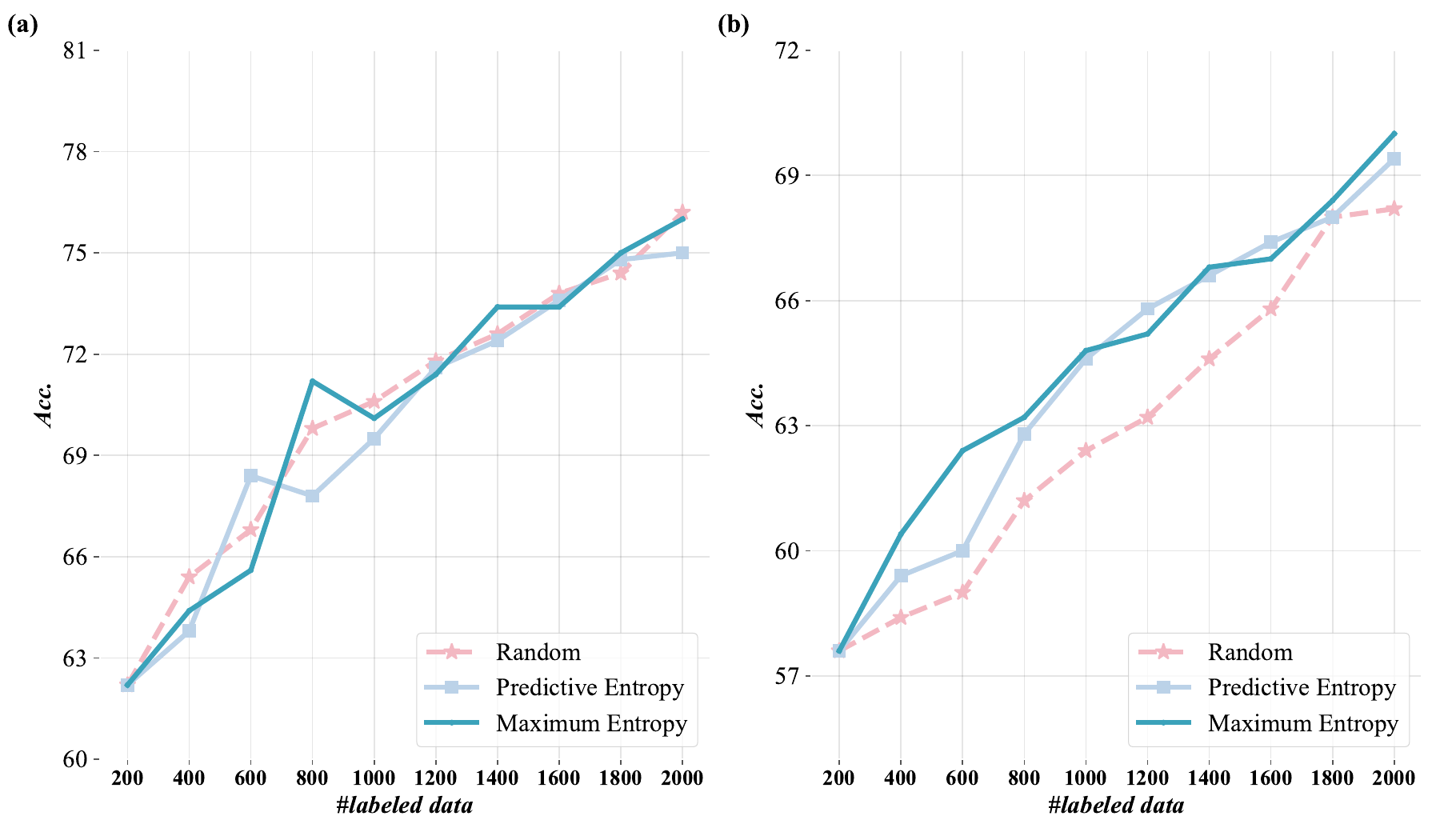}
    \caption{(a) Active learning combined with LoRA compared to passive learning. (b) Active learning combined with full parameter tuning compared to passive learning.}
    \label{fig:intro}
    \vspace{-5mm}
\end{figure}

Fine-tuning LLMs requires updating a large number of parameters, which takes a lot of time and consumes considerable memory.
Taking LLaMA-7B~\cite{touvron2023llama} as an example, fine-tuning it on a dataset of 52k instances takes over 12 hours on 4 A100 80G GPUs~\cite{Bommasani2021FoundationModels,alpaca}.
Therefore, Parameter Efficient Fine-Tuning (PEFT) methods ~\cite{houlsby2019parameter,lester-etal-2021-power,li-liang-2021-prefix,hu2021lora,ding2023parameter} and Memory Efficient Fine-Tuning (MEFT) methods~\cite{liao2023make} have been proposed. 
In addition to updating a vast number of parameters and consuming substantial memory, a neglected factor in LLMs fine-tuning is the extensive consumption of annotation data. 
Moreover, due to the inherent complexity of tasks, the human annotation resources required for fine-tuning LLMs are also significant~\cite{ouyang2022training}.

Therefore, it is important to develop the Data-Efficient Fine-Tuning (DEFT) method for LLMs. 
A common practice to improve data efficiency is active learning~\cite{cohn1996active,settles2009active}, while it has been shown that the PEFT methods can alleviate the reliance on annotated data to some extent~\cite{ding2023parameter}.
A straightforward idea for DEFT is to combine the PEFT method with active learning.
However, such a combination is not trivial. 
As shown in Figure~\ref{fig:intro} (a), simply fine-tuning an LLM with LoRA~\cite{hu2021lora} under an uncertainty-based active learning framework yields consistently inferior performances compared to passive learning (random selection of data in active learning) on the OpenBookQA dataset, while as shown in Figure~\ref{fig:intro} (b), fine-tuning LLM with full parameter under active learning yields performances better than passive learning.

To investigate the mechanism behind this uncommon phenomenon, probe experiments are conducted by investigating the prediction confidence and entropy of the LLM with LoRA during uncertainty-based active learning. 
Based on the experimental results, we deduce two potential reasons for this phenomenon.
The first issue is \textbf{\textit{uncertainty gap}}. 
To be more specific, the uncertainty calculated for selecting data during active learning comes from the full parameters, while only partial parameters remain tuned during PEFT.
It suggests that the conventional way of calculating uncertainty may not reflect the knowledge required by the PEFT parameters and therefore undermines the performance of active learning.
The second issue is \textbf{\textit{poor model calibration}} which becomes particularly significant when using the PEFT method~\cite{wang2023lora}.
It further indicates that the uncertainty calculated in the conventional way is not well-calibrated, and the data selected for active learning becomes sub-optimal.

To address the aforementioned issues, we propose 
con\textbf{S}train\textbf{T} LoRA with dynamic \textbf{A}ctive lea\textbf{R}ning~(\textbf{STAR}), a novel approach to effectively integrate uncertainty-based active learning and LoRA.
Specifically, for the uncertainty gap, we introduce a dynamic uncertainty measurement that combines the uncertainty of the base model and the uncertainty of the full model during the iteration of active learning.
For poor model calibration, we incorporate the regularization method during LoRA training to keep the model from being over-confident, and the Monte-Carlo dropout mechanism~\cite{gal2016dropout} is employed to enhance the uncertainty estimation.
Experimental results show that the proposed approach outperforms existing baseline models on three complex reasoning tasks. 
The above issues are partially resolved.

In conclusion, our contributions are three-fold:
\begin{itemize}
    \item As far as we know, we are the first to investigate and uncover the reasons why 
    directly combining active learning with LoRA fails to achieve comparable performance with passive learning through probe experiments. 
    \item A novel DEFT method, \textbf{STAR}, is proposed to effectively combine PEFT with active learning through criterion revision and model regularization.
    \item Extensive experimental results show that the proposed method addresses the issues and outperforms other baselines.
\end{itemize}

\section{Related Work}
\subsection{Efficient Fine-tuning Methods}
As LLMs continue to expand in size, the computational and financial resources required for fine-tuning these models become increasingly prohibitive.
To address this challenge, Efficient Fine-Tuning has emerged as an essential area of research~\cite{wan2023efficient}.
The methods can be classified into PEFT and MEFT~\cite{wan2023efficient,liao2023make}.
PEFT, in particular, aims to adjust a minimal subset of the model's parameters, thus conserving computational resources while maintaining or enhancing model performance~\cite{hu-etal-2023-llm}.
We classify PEFT methods into four categories:
Prompt Tuning~\cite{lester-etal-2021-power, liu2023gpt} only allow an additional $k$ tunable tokens per downstream
task to be prepended to the input text.
Prefix Tuning~\cite{li-liang-2021-prefix,liu-etal-2022-p}  keeps language model parameters frozen but optimizes a small continuous task-specific vector that pretends to be key-value pairs.
Adapter~\cite{houlsby2019parameter,he2021towards} is a new module added between layers of a pre-trained network, which is a bottleneck architecture.
Low-Rank Adaptation~(LoRA)~\cite{aghajanyan-etal-2021-intrinsic,hu2021lora} reduces the parameters and enhances computational efficiency by applying low-rank matrices.
Moreover, minimizing memory usage in fine-tuning for improving efficiency has also emerged as a critical topic~\cite{liao2023make}, with several innovative solutions being proposed.
Among these, techniques such as QLoRA~\cite{dettmers2023qlora}, QA-LoRA~\cite{xu2023qa}, and LoftQ~\cite{li2023loftq} stand out for their ability to significantly reduce memory requirements without compromising model performance.
In this paper, we focus on the application of LoRA.

\subsection{Active Learning with LLMs}
Active Learning (AL) has been extensively investigated across a multitude of NLP tasks, encompassing machine translation~\cite{miura-etal-2016-selecting,zhao-etal-2020-active}, natural language inference~\cite{snijders-etal-2023-investigating}, named entity recognition~\cite{shen-etal-2017-deep} and text classification~\cite{ein-dor-etal-2020-active,margatina-etal-2022-importance,schroder-etal-2023-small}.
In the era of LLMs, active learning is primarily employed in the selection of prompts and the annotation of data~\cite{zhang2023llmaaa,liu2023makes,xiao2023freeal}.
For instance, ~\citet{margatina-etal-2023-active} explores various active learning strategies for selecting the most relevant examples for in-context learning with LLMs.
~\citet{diao2023active} introduces an active prompting method that leverages uncertainty metrics to select questions for annotation.
In the domain of integrating PEFT with AL, \citet{jukic-snajder-2023-parameter} explored PEFT methods with different active learning in Pre-trained language models~(PLMs) such as BERT~\cite{devlin-etal-2019-bert}, demonstrating that the integration of PEFT with active learning can offer substantial performance gains.
Different from \citet{jukic-snajder-2023-parameter}, we apply decoder-only generative LLMs as the backbones, to our knowledge, we are the first to integrate LLMs combined PEFT with AL within the realm of reasoning tasks.

\section{Preliminaries}
\subsection{Parameter Efficient Fine-tuning}
Parameter-Efficient Fine-Tuning~(PEFT) methods aim to fine-tune only a small
set of external parameters while keeping the backbone model frozen and achieving comparable or even superior performance~\cite{hu-etal-2023-llm}.
The main-steam PEFT methods include the adapter-based methods~\cite{houlsby2019parameter,he2021towards}, prefix tuning~\cite{li-liang-2021-prefix}, and LoRA~\cite{hu2021lora}, among which LoRA is the most effective and widely used.
In this paper, we mainly implement PEFT with LoRA.

LoRA (\textbf{Lo}w-\textbf{R}ank \textbf{A}daptation) assumes that the updation of the model weight matrix during training is low-ranked, which can be decomposed as the multiplication of two low-rank matrices. 
\begin{equation}
    \Delta W = \alpha B A
\end{equation}
where $\Delta W$ is the updation of the model weight matrix, $B \in \mathcal{R}^{d \times r}$ and $A \in \mathcal{R}^{r \times k}$ are matrices of rank $r$, and $\alpha$ is constant scaling factor. 

During training, the model weight matrix $\bm{W}$ is fixed and only $\Delta W$ is optimized.
It is worth noticing that commonly $A$ is randomly initialized and $B$ is zero-initialized.
In this way, we have $\bm{W} = \bm{W} + \Delta W$ at the beginning of training and the fine-tuned model is identical to the base model.

\subsection{Active Learning}
Active Learning (AL) methods aim to select informative examples from the data pool to maximize the performance with the required data budget or minimize the data budget to achieve the required performance. 
The family of AL methods mainly includes uncertainty-based methods~\cite{lewis1995sequential, gal2016dropout}, diversity-based methods~\cite{sener2018active}, and discriminative-based methods~\cite{gissin2019discriminative}, where uncertainty-based methods are widely used and easy for implementation. 

In our study, we mainly consider three AL strategies, including \textsc{\small RANDOM} selection as a passive learning baseline and two uncertainty-based criteria.
Maximum Entropy~\cite{lewis1995sequential} and Predictive Entropy~\cite{duan2023shifting,kadavath2022language} are both based on uncertainty, but the former is label independent while the latter is label dependent.
The key idea behind uncertainty-based AL methods is that models will learn more efficiently from examples in which they are difficult to predict and have high prediction uncertainty.

LLMs~\cite{llama2} inherently generate sentences in a free-form and auto-regressive fashion.
This process entails the sequential prediction of the probability distribution for the subsequent token in a sentence.
Let $x$ represent the input prompt, and $s$ denote the sentence generated by the LLM, comprising $N$ tokens in total.
For any given LLM, the probability of producing a specific token $z_i$ as the $i$-th element in the sentence can be mathematically expressed as $p(z_i | s_{<i}, x)$, where $1 \leq i \leq N$.
Here, $s_{<i}$ symbolizes the sequence of previously generated tokens $\{ z_1, z_2, ..., z_{i-1}\}$.

\noindent\textbf{\textsc{\small Maximum Entropy}(\textsc{{\small ME}})} is characterized by its independence from golden response. 
It quantitatively evaluates the \textit{uncertainty} in a model's predictions by computing the entropy across all possible outcomes, formulated as:
\begin{align}
     &\textit{ME}(s, x)= \nonumber \\
  - \sum_{i=1}^{N}  &\sum_{j=1}^{V} p(v_{ij} | s_{<i}, x) \log p(v_{ij} | s_{<i}, x)
 \label{eq:me}
\end{align}
where $s$ is the generated response, $p(v_{ij} | s_{<i}, x)$ is the probability of $j$-th token in vocabulary at $i$-th element in $s$, $V$ is the vocabulary size.

\noindent\textbf{\textsc{\small Predictive Entropy}(\textsc{{\small PE})}} incorporates golden response dependency, offering a measure of the expected information gain from the true label, given the predictive distribution.
It is formulated as:
\begin{align}
    \textit{PE}(\overline{s}, x) &= - \log p(\overline{s}|x)  \nonumber \\
    &= \sum_{i=1}^{N}{-\log p(\overline{z}_i|\overline{s}_{< i}, x)}
\label{eq:pe}
\end{align}
where $\overline{s}$ is the golden response, $p(\overline{z}_i | \overline{s}_{<i}, x)$ is the probability of $i$-th token in the golden response.

\section{Probing PEFT on Prediction Uncertainty}
In this section, we describe how to design probe experiments to investigate the reason behind the failure of LoRA combined with AL methods.
We will first introduce the experiment setup, and then we will talk about how to prob LoRA under the AL framework with prediction confidence and prediction entropy. 
We also discuss how to conclude from the experimental results.

\subsection{Probe Experiment Design}
As uncertainty-based AL methods mainly depend on the confidence or uncertainty of model predictions to select examples during each iteration, it is straightforward to probe the confidence and uncertainty of model predictions during AL iterations.

We mainly focus on two training variants of LLMs. 
The first one, denoted as \textit{PEFT} method, is LLM finetuned with LoRA, which is shown to be problematic under the AL framework.
The second one, denoted as \textit{Few-shot} method, is untuned LLM, which is taken as a control group.
To enhance the performance of untuned LLM on downstream tasks, we employ In-Context Learning~\cite{dong2022survey} prompting by adding demonstrations with the input prompt.
LLaMA-2~\cite{llama2} serves as the backbone LLM. 
Experiments are conducted on the BoolQ dataset~\cite{clark-etal-2019-boolq} because its labels only include ``\textit{true}'' and ``\textit{false}'', which makes the prediction uncertainty and prediction confidence easy to calculate.

\subsection{Probing with Prediction Confidence}
\label{sec:over-confidence}

\begin{figure}[t]
    \centering
    \includegraphics[width=0.45\textwidth]{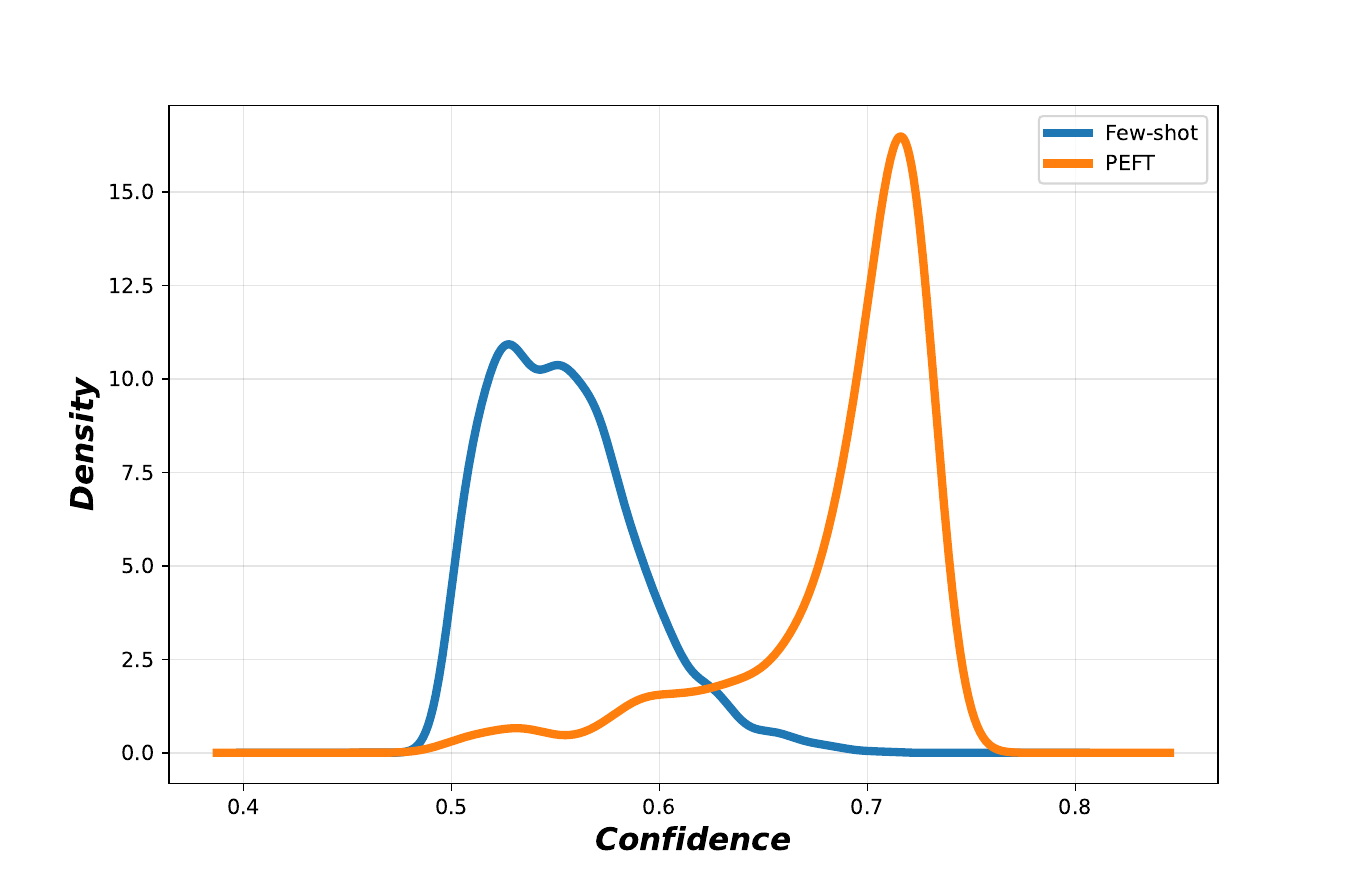}
    \caption{Density plot of confidence for wrong predictions.}
    \label{fig:density}
\end{figure}

The first probe experiment is designed to explore whether the model prediction confidence of the \textit{PEFT} method exhibits issues compared to \textit{Few-shot} methods.
The prediction confidence $CF$ is measured by the maximum between the output probability on token ``\textit{true}'' and ``\textit{false}''.
\begin{equation}
    CF = max(p_{true},p_{false})
\end{equation}
where $p_{true}$ and $p_{false}$ denotes the probabilities of token ``\textit{true}'' and ``\textit{false}'', respectively.

Then the prediction confidence of \textit{PEFT} and \textit{Few-shot} are calculated and the density plot is drawn to make a comparison between these two methods as shown in Figure~\ref{fig:density}.
To mitigate differences in model accuracy, we only consider the confidence of the model for the \textbf{wrong} predictions on the test set of BoolQ.
The intuition is that for examples that the model is less likely to predict right, they should be less confident.

The \textit{PEFT} method achieves an accuracy of 73.36\%, which is much higher than the \textit{Few-shot} method with an accuracy of 45.41\%. 
As shown in Figure~\ref{fig:density}, the \textit{PEFT} method is overconfident compared to the \textit{Few-shot} method, where the confidence of the wrong prediction is as high as 70\%, which indicates a \textbf{\textit{model calibration issue}}.

\subsection{Probing with Prediction Entropy}

\begin{figure*}[t]
    \centering
    \includegraphics[width=0.9\textwidth]{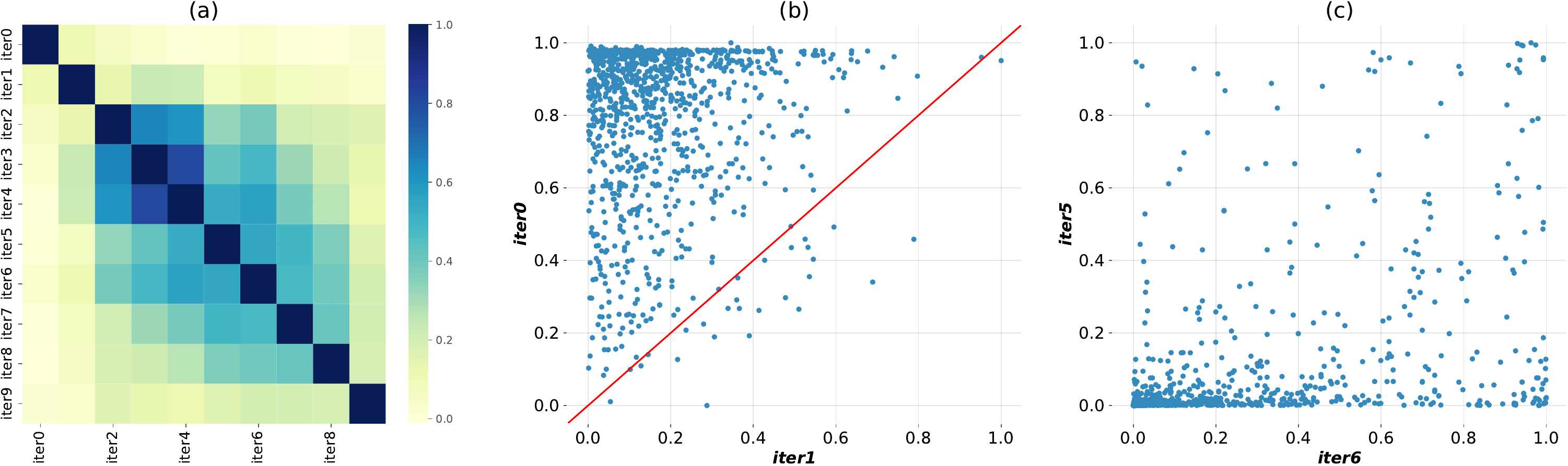}
    \caption{(a) Heatmap of correlation between prediction entropy across different iterations;
    (b) Scatter plot for prediction entropy between base model (Iter0) and model after first iteration (Iter1);
    (c) Same as (b), except values are taken from Iter5 and Iter6.}
    \label{fig:entropy}
\end{figure*}

The second probe experiment is designed to investigate the change of prediction entropy of \textit{PEFT} model during active learning iteration.
The \textsc{\small Maximum Entropy}(\textsc{{\small ME}}) is employed as the uncertainty during active learning.
Nine rounds of iteration are performed with 500 examples selected during each iteration.
The \textit{PEFT} model is trained with 500 examples at the beginning as a warm-up.

The correlation between examples with top 1000 entropy at the beginning is calculated and the heatmap of the correlation is shown in Figure~\ref{fig:entropy} (a).
As we can observe in Figure~\ref{fig:entropy} (a), the correlation between the base model (model without PEFT tuning) and models after AL iteration is close to 0, which indicates \textbf{\textit{a clear gap between the base model and \textit{PEFT} model}}.
This phenomenon is even clear with the scatter plot in Figure~\ref{fig:entropy}, where the dots in Figure~\ref{fig:entropy} (b) should appear around the red line but appear in the upper triangular region.
In Figure~\ref{fig:entropy} (c), the correlation coefficients of entropy between the two iterations become relatively normal, which is consistent with Figure~\ref{fig:entropy} (a), suggesting that the gap between iterations has been alleviated.

\section{Methods}
\label{sec:methods}
In this section, we introduce the proposed method \textbf{STAR} in detail. 
We will first describe the overall workflow of \textbf{STAR},
then we will discuss methods to address the \textbf{\textit{uncertainty gap}} issue and the \textbf{\textit{model calibration}} issue.
Finally, we will conclude the proposed method with a pseudocode.

\begin{figure*}[t]
    \centering
    \includegraphics[width=0.95\textwidth]{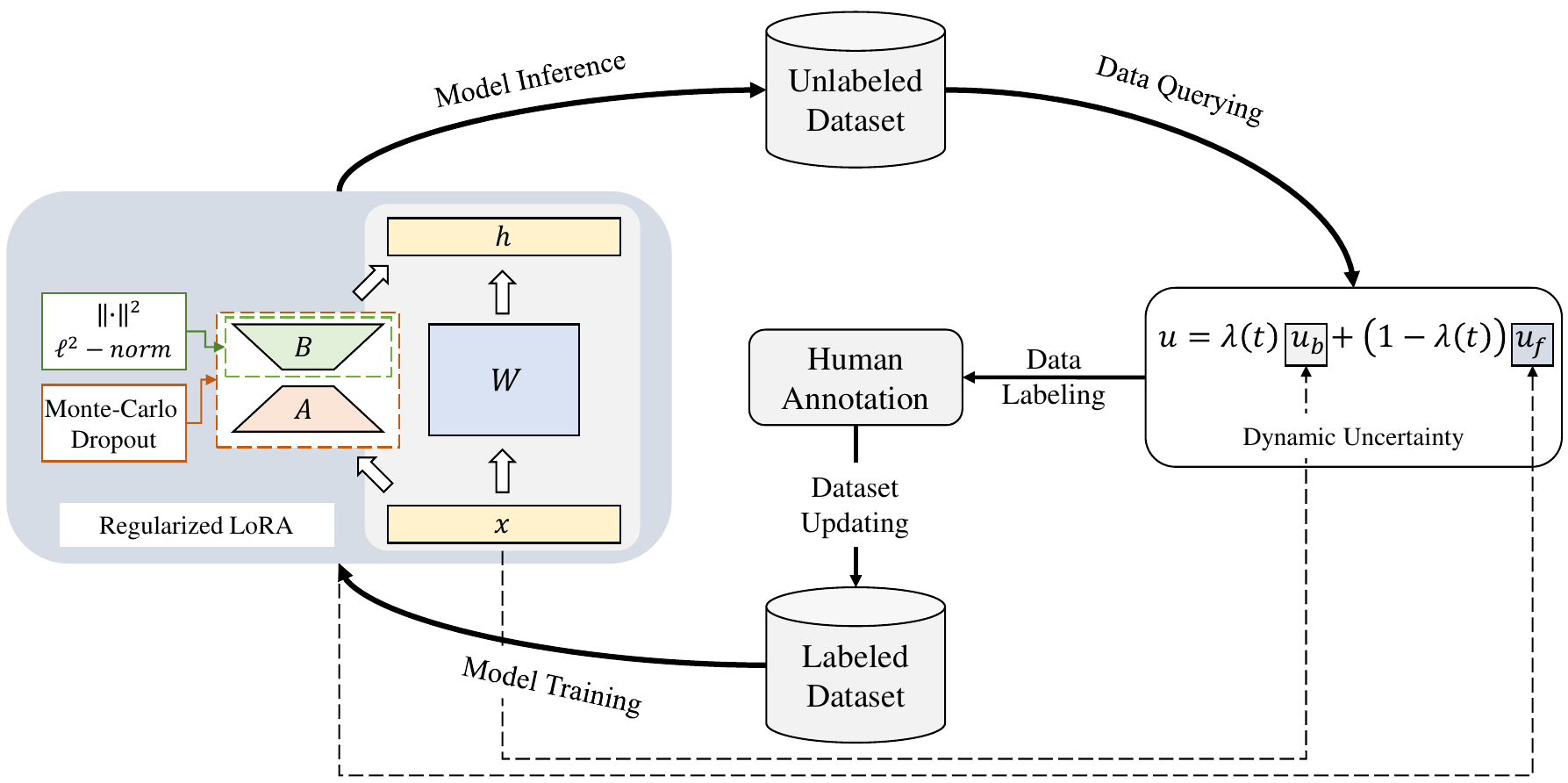}
    \caption{The framework of \textbf{STAR}.
    It primarily consists of five steps: Model Inference, Data Querying, Data Labeling, Dataset Updating, and Model Training.}
    \label{fig:framework}
\end{figure*}

\subsection{LoRA under Active Learning Iteration}
As shown in Figure~\ref{fig:framework}, the $k$-th iteration of \textbf{STAR} consists of the following steps.
\begin{enumerate}[]
\item \textbf{Model Inference} that employs the present model $M_k$ to make inference on unlabeled dataset $D_k^{U}$.
\item \textbf{Data Querying} that selects the most informative examples to form a subset $S_k^{U}$ with the results of inference based on the dynamic uncertainty estimation method.
\item \textbf{Data Labeling} that labels the unlabeled subset $S_k^{U}$ to form the labeled subset $S_k^{L}$.
\item \textbf{Dataset Updating} that updates the labeled dataset $D_k^{L}$ by appending appending the labeled subset $D_{k+1}^{L} = D_k^{L} \cup S_k^{L}$. 
\item \textbf{Model Training} that updates the present model with new labeled dataset $D_{k+1}^{L}$ to get model $M_{k+1}$ for next iteration. 
\end{enumerate}

\subsection{Dynamic Uncertainty Measurement}
To address the issue of \textbf{\textit{uncertainty gap}}, we proposed a dynamic uncertainty measurement to integrate the uncertainty of the frozen LLM~(base model) and the uncertainty of LLM fine-tuned with LoRA~(full model) dynamically based on the AL iteration.

The key idea is that at the beginning of PEFT training, the extra parameters are under-fitting, where the uncertainty calculated is less reliable than the frozen parameters.
As the iteration of active learning increases, the uncertainty of the full model becomes more reliable, which is similar to the zero-initialized attention weight in LLaMA-adapater~\cite{zhang2023llama}.
\begin{equation}
    \mu = \lambda(t) \mu_b + (1-\lambda(t)) \mu_f
\end{equation}
where $\mu_b$ and $\mu_f$ denote the prediction uncertainty of the base model and the full model respectively, $\lambda(t) \in [0,1]$ is a monotone decreasing function of AL iteration $t$.
Note that, our measurement approach only requires one additional computation of the base model at the beginning, which remains constant throughout and does not significantly increase the FLOPs.

\subsection{Calibration with Hybrid Regularization}
To address the issue of \textbf{\textit{poor model calibration}}, we propose a hybrid regularization method during PEFT training.
As discussed in Section~\ref{sec:over-confidence}, the PEFT model demonstrates a pronounced tendency toward over-confidence, which indicates that the model is over-fitting. 

Common approaches to prevent the model from being over-fitting include early-stoping~\cite{doan2004generalization}, regularizations~\cite{santos2022avoiding}, and ensemble methods~\cite{ganaie2022ensemble}. 
Considering the difference between LoRA parameters $A$ and $B$, we integrate two regularization methods into a hybrid regularization to keep LoRA from being over-fitting.

For the $B$ matrix, which is zero-initialized, a $L^2$ norm weight decay is employed.
\begin{equation}
    B_t \leftarrow B_{t-1} - \gamma (g_{t-1} - \beta B_{t-1})
\label{eq:l2norm}
\end{equation}
where $g_{t-1}$ denotes the normalized gradient acquired from the standard Adam optimizer, and $\beta$ denotes the strength of regularization.

For the $A$ matrix, which is randomly Gaussian initialized $N(0,1)$, the Monte-Carlo dropout (MC dropout)~\cite{gal2016dropout} is adopted for more robust uncertainty estimation.
MC dropout works by activating the dropout unit both in the training and inference stages, which can be regarded as an approximation to the Bayesian Neural Network.
With the dropout unit activated during the inference stage, neural networks can generate different outputs with the same input, where expectations can be taken for more robust estimation. 
\begin{equation}
\begin{split}
    \mu_f &= \frac{1}{K}\sum_k \mu_f^{(k)} \\
    \mu_f^{(k)} &= \textit{ME}(\text{LLM}(x|\hat{A}_k,\hat{B}_k)) 
\end{split}
\label{eq:mcdropout}
\end{equation}
where $K$ denotes the number of feedforward propagations during the inference stage, $\mu_f^{(k)}$ denotes the uncertainty estimated at $k$-th feedforward, $\hat{A}_k$ and $\hat{B}_k$ denote the LoRA matrices sampled from $A$ and $B$ with dropout unit activated.

\subsection{Overall Algorithm}
The overall algorithm of \textbf{STAR} is shown in Algorithm~\ref{alg}.

\begin{algorithm}[h]
\caption{STAR}
\textbf{Input}:unlabeled dataset $D^{U}$, labeled dataset $D^{L}$,
the LLM $M$, number of iteration $N$, size of subdataset during iteration $m$.
\begin{algorithmic}[1]
    \STATE initialize $D_0^{U}$ and $D_0^{U}$
    \STATE warm-up LLM $M_0$ 
    \STATE \textbf{for} $k=0$ \textbf{to} $N$:
    \STATE \quad making inference with $M_k$ on $D_k^{U}$
    \STATE \quad querying subset $S_k^{U}$ from $D_k^{U}$ based on Equation~\eqref{eq:me}  
    \STATE \quad updating $D_{k+1}^{U} \leftarrow D_{k}^{U} \setminus S_k^{U}$ 
    \STATE \quad labeling $S_k^{U}$ to get $S_k^{L}$
    \STATE \quad updating $D_{k+1}^{L} \leftarrow D_{k}^{L} \cup S_k^{L}$ 
    \STATE \quad fine-tuning LLM $M_k$ to get $M_{k+1}$ on $D_{k+1}^{L}$ based on Equation~\eqref{eq:l2norm} and Equation~\eqref{eq:mcdropout} 
    \RETURN LLM after fine-tuning $M_N$
\end{algorithmic}
\label{alg}
\end{algorithm}

\begin{table*}[]
\small
\centering
\begin{tabular}{lcccccc}
\toprule
Method & \multicolumn{2}{c}{\textbf{GSM8K}}    & \multicolumn{2}{c}{\textbf{BoolQ}} & \multicolumn{2}{c}{\textbf{OpenBookQA}}  \\
\midrule
&\texttt{AUC}&\texttt{RIPL}&\texttt{AUC}&\texttt{RIPL}&\texttt{AUC}&\texttt{RIPL}\\
\midrule
\textsc{\small Random}  &  27.37 & -& 60.46  &- & 63.44 &-\\
\midrule
\textsc{\small Predictive Entropy}  & 27.30 & -0.09 & 58.39 & -5.24 &63.05&-1.07\\
\quad  w/~\textbf{STAR} & \underline{28.40} & \underline{1.42} 
&\underline{61.84} & \underline{3.49}&\underline{64.86} &\underline{3.88}\\
\midrule
\textsc{\small Maximum Entropy}   & 27.16 & -0.28 & 60.65 & 0.48& 63.36&-0.22  \\
\quad w/~\textbf{STAR}  &  \textbf{28.83} & \textbf{2.01} & \textbf{61.91} & \textbf{3.67}& \textbf{66.17} &\textbf{7.47}\\
\bottomrule
\end{tabular}
\caption{The performance of different methods in a passive learning setup in terms of the \texttt{AUC} and \texttt{RIPL}.
The optimal results among all methods are \textbf{bolded} and the second-best results are \underline{underlined}.}
\label{tab:main_result}
\end{table*}

\begin{figure*}[ht]
    \centering
    \includegraphics[width=0.8\textwidth]{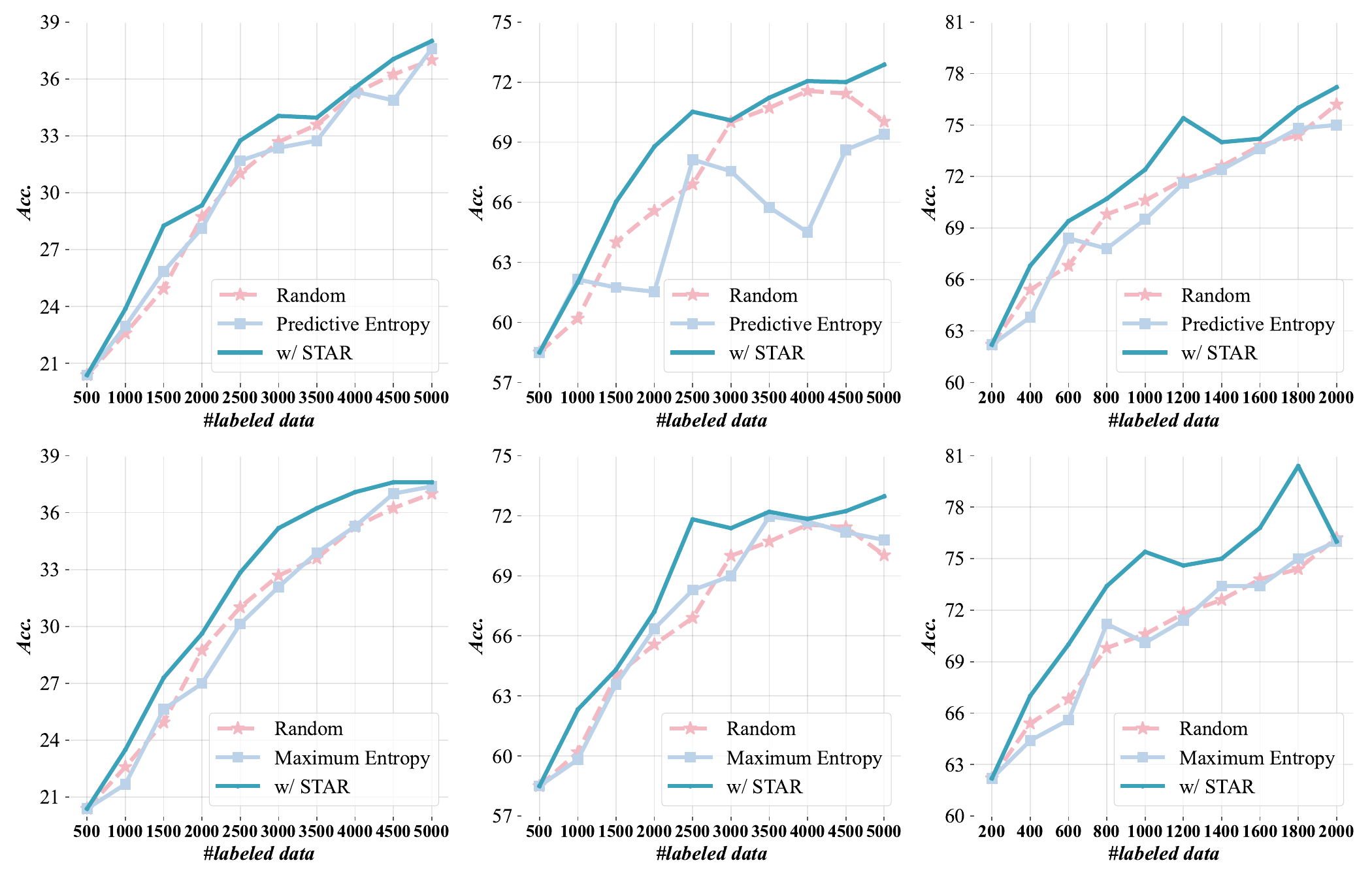}
    \caption{
    The Learning curves comparing the  \textsc{\small Predictive Entropy} and \textsc{\small Maximum Entropy} methods, and each w/ \textbf{STAR}, against the \textsc{\small RANDOM} baseline.
    The first column corresponds to the GSM8K dataset, the second column to the BoolQ dataset, and the third column to the OpenBoolQA dataset.}
    \label{fig:curve}
    \vspace{-1.0em}
\end{figure*}

\section{Experiments}
\subsection{Datasets}
In our research, we employ three benchmark datasets spanning two categories of reasoning problems for AL evaluation:\\
\textbf{Arithmetic Reasoning}: the GSM8K dataset~\cite{cobbe2021gsm8k} comprises approximately 8.5K high-quality linguistically diverse grade school math word problems created by human problem writers.\\
\textbf{Commonsense Reasoning}:
(1) the BoolQ dataset~\cite{clark-etal-2019-boolq} is a specialized question-answering dataset designed for yes/no questions; 
(2) the OpenBookQA dataset~\cite{mihaylov-etal-2018-suit} is a four-way multiple-choice question-answering dataset.

See Appendix~\ref{appendix:prompt} for more details about the dataset.

\subsection{Settings}
\noindent\textbf{Experimental setup}
In the experiment conducted on the GSM8K and BoolQ datasets, we incrementally selected 500 new instances in each step of the AL experiment.
The initial warm start for the AL setting is established by randomly choosing 500 instances.
Furthermore, we adhere to a labeling budget constraint of 5,000 instances for each dataset.
Considering the size of the training set for OpenBookQA, we design the AL framework to incrementally select 200 new instances during each iteration.
The labeling budget for this process is set to 2,000 instances.
The details regarding the evaluation can be found in Appendix~\ref{sec:eval}.

\noindent\textbf{Implementations}
In the empirical study, we utilize the state-of-the-art openly accessible LLM, LLaMA2-7B~\cite{llama2}\footnote{\url{https://huggingface.co/meta-llama/Llama-2-7b}} as the base model.
For comprehensive details on the hyperparameters employed in our experiments, please refer to Appendix~\ref{appendix:para}.

\begin{table*}[th]
\small
\centering
\begin{tabular}{lcccccc}
\toprule
Method & \multicolumn{2}{c}{\textbf{GSM8K}}    & \multicolumn{2}{c}{\textbf{BoolQ}} & \multicolumn{2}{c}{\textbf{OpenBookQA}}  \\
\midrule
&\texttt{AUC}&\texttt{RIPL}&\texttt{AUC}&\texttt{RIPL}&\texttt{AUC}&\texttt{RIPL}\\
\midrule
\textsc{\small Predictive Entropy} & 27.30 & -0.09 & 58.39 & -5.24& 63.05&-1.07  \\
\quad +Dynamic & 27.73 & 0.59 & 60.80 & 0.86& 63.81&1.01  \\
\quad \quad +Monte-Carlo dropout & 28.04 & 1.02 & 61.30 & 2.12& 64.16&1.97  \\
\quad \quad + $L^2$ norm weight decay & 27.93 & 0.87 & 60.99 & 1.34& 64.13&1.89  \\
\midrule
\textsc{\small Maximum Entropy} & 27.16 & -0.28 & 60.65 & 0.48& 63.36&-0.22  \\
\quad +Dynamic & 27.68 & 0.52 & 61.01 & 1.39 & 64.51&2.95  \\
\quad \quad +Monte-Carlo dropout& 28.03 & 1.00 & 61.72 & 3.19& 65.17&4.73  \\
\quad \quad + $L^2$ norm weight decay& 27.82 & 0.72 & 61.29 & 2.10 & 64.72&3.50 \\
\midrule
\end{tabular}
\caption{The ablation performance of different methods,  \texttt{AUC} and \texttt{RIPL} are reported.}
\label{table:ablation}
\end{table*}

\section{Result and analysis}
\subsection{Main Result}
Table~\ref{tab:main_result} presents a detailed comparison of different methods' performance, evaluated across three different datasets: GSM8K, BoolQ, and OpenBookQA.  
\textsc{\small Random} serves as a fundamental baseline, with its \texttt{AUC} listed.

Both original \textsc{\small PE} and \textsc{\small ME} methods underperform compared to \textsc{\small Random} on these three datasets in terms of \texttt{AUC}.
The \texttt{RIPL} metric also hovers around zero, indicating that the original AL strategy is essentially ineffective.

After applying our proposed \textbf{STAR} method, \textsc{\small PE} and \textsc{\small ME} exhibit superior performance across all datasets and metrics.
For instance, in the GSM8k dataset, \textsc{\small ME} w/~STAR achieves an \texttt{AUC} of \textbf{28.83} and a \texttt{RIPL} of \textbf{2.01}, indicating a notable advancement over the baseline \textsc{\small Random} and \textsc{\small ME}.
The improvements are most pronounced in the OpenBookQA dataset, where \textsc{\small ME} w/~STAR method achieves a remarkable \texttt{RIPL} of \textbf{7.47}.
Furthermore, in the BoolQ dataset, \textsc{\small ME} w/~STAR achieves higher performance compared to the \textsc{\small PE} w/~STAR.
This pattern of \textsc{\small ME} w/~STAR outperforming \textsc{\small PE} w/~STAR is consistent across the GSM8K and OpenBookQA datasets as well.
These results suggest that the \textsc{\small ME} w/~STAR is more effective. 

Then, we explore how the models’ performance changes as the training set increases.
Figure~\ref{fig:curve} shows the learning curves for corresponding AL methods on GSM8K, BoolQ, and OpenBookQA datasets, respectively.
The \textsc{\small Random} baseline and the two original active learning approaches perform comparably, suggesting that the active learning methods appear to be ineffective.
Notably, the BoolQ dataset exhibits particularly high variability in results when using the \textsc{\small PE}  strategy, which may be attributed to BoolQ's binary output format of ``\textit{true}'' and ``\textit{false}''.
The gap between the full model and the base model could easily lead to skewed predicting results in a single iteration.

It is evident that w/ STAR methods demonstrate the most significant improvement on the OpenBookQA dataset.
After applying our method, the model learned truly more useful samples. 
For instance, as evidenced by the BoolQ dataset in Figure~\ref{fig:curve}, the performance of the model reaches saturation with just 1000 samples.
This indicates that the selected samples are sufficiently diverse and useful for model learning.

\subsection{Ablation Study}
Since our methods have two main components, which are dynamic uncertain measurement and calibration with hybrid regularization as described in Sec~\ref{sec:methods}. We conduct a detailed ablation study to assess the effect of the two components.
As shown in Table~\ref{table:ablation}, upon employing the dynamic uncertain measurement, all \texttt{AUC} are improved, and the \texttt{RIPL} turns positive.
This indicates a significant gap between the full model and the base model in the original strategies, which our dynamic indicator effectively mitigated.

Subsequently, building on this foundation and individually incorporating MC dropout and $L^2$ norm weight decay,
it is observed that both constraint methods enhance performance, with MC dropout offering a more substantial improvement.
The addition of calibration methods indeed effectively mitigates the issue of model over-confidence and improves model calibration.

\section{Conclusion}
In this paper, to improve the data efficiency of Large Language Models~(LLMs) during the fine-tuning process, we propose a data-efficient parameter tuning method by combining LoRA with active learning. 
To address the issue that uncertainty-based active learning fails to combine with LoRA, we experimentally identify and summarize two possible reasons: uncertainty gap and poor model calibration.
To resolve the uncertainty gap issue, we propose a dynamic uncertainty calculation method, and to address poor model calibration, we introduce a regularization-based constraint method. 
By integrating these two approaches, we partially solve the aforementioned failure issues. 
Extensive experiments show that our proposed method outperforms baseline models on multiple reasoning datasets.

\section*{Limitations}
Though achieving promising results in the experiments, our work still has the following limitations.
\begin{itemize}[leftmargin=*]
    \item Due to constraints on computational resources, we did not conduct experiments on larger versions of LLaMA2 from 13B to 70B, nor did we experiment with other types of LLMs including BLOOM, Falcon, $\etc$ 
    \item Due to limitations in computational resources and time, we did not explore the combination of other types of PEFT methods (series/parallel adapters, prefix tuning) with different types of active learning methods (diversity-based active learning). 
    Therefore, the validity of the methods and conclusions in this paper for a wider combination of PEFT and active learning remains unexplored. 
    Further work should include exploring a more extensive combination of PEFT and active learning.
    \item We only speculated on the reasons for the failure of combining LoRA with active learning through simple probe experiments, without delving deeper into the underlying mechanisms. 
    Future work should involve exploring the deeper mechanisms behind this phenomenon.
\end{itemize}

\section*{Acknowledgement}
The authors would like to thank the anonymous reviewers for their insightful comments. This work is funded by the National Natural Science Foundation of China (62176053). This work is supported by the Big Data Computing Center of Southeast University.

\bibliography{anthology,custom}
\bibliographystyle{acl_natbib}
\newpage
\appendix

\section{Dataset}
\label{appendix:prompt}
Table~\ref{table:sta} shows the statistics of the dataset.
In light of the unique tasks associated with each dataset, we implement a structured template approach.
This template tailors the content and responses to the specificities of each dataset.
We give the data templates for each dataset used to fine-tune LLM in Table~\ref{table:prompt}.
\begin{table}[!t]
	\centering
	\small
\begin{tabular}{lccc}
\toprule
Dataset & \textbf{\#train} & \textbf{\#test} & \textbf{Answer Format} \\
\midrule
GSM8K  &    7,473     &   1,319     &     Number   \\
BoolQ   &   9,427      &    3,270    &    Letter    \\
OpenBookQA    &   4,957      &   500     &   Letter   \\
\bottomrule 
\end{tabular}
\caption{Details of datasets being evaluated.}
\label{table:sta}
\end{table}

\begin{table*}[!t]\centering
\small
\begin{tabular}{lp{11cm}}\hline
Dataset & Fine-tuning Data Template \\
\hline
GSM8K & [QUESTION] 

Answer the above question. First, think step by step and then answer the final number.

[ANSWER] \\ \hline
BoolQ & [QUESTION]

The correct answer is

[ANSWER] \\ \hline
OpenBookQA & [QUESTION] 

Answer1: [ANSWER\_1]

Answer2: [ANSWER\_2]

Answer3: [ANSWER\_3]

Answer4: [ANSWER\_4]

The correct answer is 
[ANSWER]  \\
\hline
\end{tabular}
\caption{The fine-tuning data template for each dataset.}
\label{table:prompt}
\end{table*}

\section{Model Hyper Parameters}
\label{appendix:para}
Following the prior works~\cite{hu2021lora, li2023loftq}, we maintain the original weights of the backbone architecture unchanged and integrate low-rank adapters into the Multi-Head Attention(MHA) and Feed-Forward Network(FFN) components of all layers.
These low-rank adapters are configured with a rank of 64 and a factor of $\alpha$ set to 16, alongside a dropout rate of 0.1 to mitigate overfitting.
The model parameters are optimized by AdamW~\cite{loshchilov2018decoupled}.
We use a batch size of 8 and a learning rate of 1.5e-4 for the GSM8K task and a batch size of 32 and a learning rate of 3e-5 for the BoolQ task and the OBQA task.
In the AL setting, the model is trained for a fixed number of epochs: 3 epochs for the GSM8K task, and 15 epochs for both the BoolQ and OBQA tasks.
All reported results are averaged over three runs.
Our implementation leverages the \textit{PyTorch}\footnote{\url{https://github.com/pytorch/pytorch}} framework and \textit{HuggingFace Transformers}\footnote{\url{https://github.com/huggingface/transformers}} library~\cite{wolf-etal-2020-transformers}.
Our experiments are carried out with an NVIDIA A100 80GB GPU.

\section{Evaluation}
\label{sec:eval}
Following previous work~\cite{schroder-etal-2022-revisiting, jukic-snajder-2023-smooth, jukic-snajder-2023-parameter}, our study utilizes the Area Under the Curve (\texttt{AUC}) metric to assess the comprehensive efficacy of the methods we propose. 
The accuracy metric (\textit{Acc.}) is employed for evaluating the effectiveness at each individual AL step. 

To ascertain the success of AL, we compute the Relative Improvement over Passive Learning (\texttt{RIPL}), delineated as follows:
\begin{equation}
    RIPL(S_{AL}, S_{PL}) = \frac{AUC(S_{AL})-AUC(S_{PL})}{1-AUC(S_{PL})}
\end{equation}
where $S_{AL}$ and $S_{PL}$  denotes $AL$ methods and \textsc{\small RANDOM} method.
RIPL serves as an estimator for the quotient of the maximal attainable enhancement that an AL approach can secure over the conventional passive learning benchmark.
A \texttt{RIPL} score of $1$ signifies the epitome of theoretical enhancement, equating to achieving an \textit{Acc.} of $1$ during the initial sampling phase and maintaining this optimum performance across all subsequent stages.
In contrast, a \texttt{RIPL} score below $0$ suggests that the AL strategy is outperformed by passive learning approaches.
\end{document}